\documentclass[10pt,twocolumn,letterpaper]{article}

\usepackage{iccv}      

\usepackage{graphicx}	
\usepackage{amsmath}	
\usepackage{amssymb}	
\usepackage{booktabs}
\usepackage{times}
\usepackage{microtype}
\usepackage{epsfig}
\usepackage{caption}
\usepackage{float}
\usepackage{placeins}
\usepackage{color, colortbl}
\usepackage{stfloats}
\usepackage{enumitem}
\usepackage{tabularx}
\usepackage{xstring}
\usepackage{multirow}
\usepackage{xspace}
\usepackage{url}
\usepackage{subcaption}
\usepackage{xcolor}
\usepackage[hang,flushmargin]{footmisc}

\definecolor{iccvblue}{rgb}{0.21,0.49,0.74}
\usepackage[pagebackref,breaklinks,colorlinks,allcolors=iccvblue]{hyperref}

\def\paperID{*****} 
\def\confName{ICCV}
\def\confYear{2025}

\title{Measuring Image-Relation Alignment: Reference-Free Evaluation of VLMs and Synthetic Pre-training for Open-Vocabulary Scene Graph Generation}

\author{Maëlic Neau$^{1}$ \hspace{0.3in} Zoe Falomir $^1$ \hspace{0.3in} Cédric Buche$^{2,3}$ \hspace{0.3in} Akihiro Sugimoto$^{4}$\\
$^1$Computing Science Department, Umeå University, Sweden\\
$^2$CNRS IRL 2010 CROSSING, Australia\\
$^3$IMT Atlantique, France\\
$^4$National Institute of Informatics, Japan
\\{\tt\small \{maelic.neau,zoe.falomir\}@umu.se,
cedric.buche@cnrs.fr, sugimoto@nii.ac.jp}
}

\begin{document}

\maketitle

\begin{abstract}

Scene Graph Generation (SGG) encodes visual relationships between objects in images as graph structures. Thanks to the advances of Vision-Language Models (VLMs), the task of Open-Vocabulary SGG has been recently proposed where models are evaluated on their functionality to learn a wide and diverse range of relations. Current benchmarks in SGG, however, possess a very limited vocabulary, making the evaluation of open-source models inefficient. In this paper, we propose a new reference-free metric to fairly evaluate the open-vocabulary capabilities of VLMs for relation prediction. Another limitation of Open-Vocabulary SGG is the reliance on weakly supervised data of poor quality for pre-training. We also propose a new solution for quickly generating high-quality synthetic data through region-specific prompt tuning of VLMs. Experimental results show that pre-training with this new data split can benefit the generalization capabilities of Open-Voc SGG models\footnote{Code and data available at \url{https://github.com/Maelic/OpenVocSGG}}.

\end{abstract}
\section{Introduction}
\label{sec:intro}

The task of Scene Graph Generation (SGG) has gained growing interest in recent years. SGG aims at generating a structured abstract representation of image content as a directed acyclic graph composed of $<subject, predicate, object>$ triplets, where each $subject$ and $object$ is traditionally grounded to an image region through 2D coordinates (bounding boxes). 
Applications of such representations include Visual-Question Answering (VQA) \cite{schaferVisualAnalysisSceneGraphBased2023}, Image Captioning \cite{yangTransformingVisualScene2023}, or even Robotics \cite{neauDefenseSceneGraph2024}. While a considerable amount of work has tackled the task of SGG in a closed-vocabulary fashion \cite{zellersNeuralMotifsScene2018a,tangUnbiasedSceneGraph2020b,yangPanopticSceneGraph2022}, approaches targeting the task in open-vocabulary settings are scarce \cite{chenExpandingSceneGraph2025,heOpenVocabularySceneGraph2022}. Open-Vocabulary SGG (OV-SGG) is a critical task to enable a new range of applications in robotics \cite{amodeoOGSGGOntologyGuidedScene2022} or embodied agent perception \cite{liEmbodiedSemanticScene2022a}, where high-quality annotated data can be difficult to acquire.

Similar to other Open-Vocabulary tasks \cite{gupta2019lvis}, OV-SGG approaches are evaluated on closed-vocabulary benchmarks (Visual Genome \cite{krishna2017visual}, PSG \cite{yangPanopticSceneGraph2022}) in a zero-shot manner. However, current benchmarks only evaluate models on a small number of relations (i.e., $< 50$), which limits our understanding of the generalization capabilities of such models. This paper aims to extend the metrics used in OV-SGG with the introduction of a reference-free evaluation to grade the efficiency of models where there is a lack of groundtruth for data annotations.

OV-SGG approaches often rely on a large amount of data for pre-training. Recent work proposes to generate pre-training data for OV-SGG through caption-image regions alignment \cite{yuanRLIPv2FastScaling2023} or directly through the use of state-of-the-art Vision-Language Models (VLMs) \cite{wangAllSeeingProjectV22025}. This paper shows that the former approaches results in low-quality coarse annotations, which may hinder the pre-training of OV-SGG models. Regarding the latter approach, the current literature is missing a comprehensive review of the abilities of VLMs to generate high-quality, consistent, and diverse visual relations. This paper proposes an extensive investigation of the performance of different VLMs for the prediction of fine-grained visual relationships. We apply the results of our investigation to generate high-quality synthetic relationship annotations for 200K images. We call this new dataset the Fine-Grained Open-Vocabulary SGG dataset (FG-OV SGG). In a series of experiments, we show that our generated data is of better quality than the existing previous data in the literature.

This paper describes the following contributions:
\begin{enumerate}
    \item A new reference-free metric, the RelCLIPScore, to evaluate OV-SGG models and VLMs in unconstrained settings (Section \ref{sec:new-metrics});
    \item A quantitave evaluation on the VLM performance for region-specific relation prediction, highlighting their strengths and weaknesses for the task (Section \ref{sec:evaluation}); and
    \item A qualitative evaluation providing a new dataset, FG-OV SGG, generated using VLMs evaluated using RelCLIPScore metrics, which can be beneficial to Open-Vocabulary SGG models (Section \ref{sec:FG-OV}). 
\end{enumerate}


\section{Related Work}
\label{sec:related}

This section presents the most recent works regarding open-vocabulary SGG metrics and weakly annotated data.

\textbf{Open-Vocabulary SGG Metrics.} Traditional approaches evaluate Open-Vocabulary Relationship Detection and Open-Vocabulary Scene Graph Generation using mean Average Precision (mAP) \cite{yuanRLIPv2FastScaling2023} or Recall@K/meanRecall@K \cite{heOpenVocabularySceneGraph2022,kimLLM4SGGLargeLanguage2024} metrics on standard SGG benchmarks \cite{krishna2017visual,yangPanopticSceneGraph2022} in a zero-shot manner. The Recall@K evaluates the performance of a model on the top predictions (i.e. top 20/50/100) for each image, independent of the class. The meanRecall@K evaluates the performance on the average of all predicate classes, which is more significant for long-tail learning such as in the task of SGG.  However, using ground truth references to evaluate Open-Vocabulary models is not optimal for the following reasons:
\begin{enumerate}[i.]
    \item Relations-based datasets are not extensively annotated, leaving a considerable amount of true relations unannotated (i.e. miss positive samples \cite{zhangFineGrainedSceneGraph2022})
    \item The SGG task includes a multi-label classification paradigm since multiple predicates can be correct for a given $<subject, object>$ pair \cite{tangUnbiasedSceneGraph2020b}.
    \item Evaluating the task of SGG with ranking metrics does not give a good overview of the model performance. For instance, a true positive predicted at rank 1 and 99 will lead to the exact same Recall@100 score.
\end{enumerate}

In a recent work \cite{chenWhatMakesScene2025}, the SGScore has been proposed to evaluate the alignment of Image Generation models with complex scenes. The SGScore computes the similarity between an image and a graph by asking multiple-choice questions to an "oracle" VLM (such as: \textit{What is the relationship between the person and the sports ball in the image? A) kicking; B) throwing; C) holding; D) no visible relationship.} \cite{chenWhatMakesScene2025}). The authors of SGScore do not give any details on the choice of negative examples for the question-answer pairs which hindered reproducibility.
Also, multiple-choice questions can not account for multi-label instances, which are overly present in Open-Vocabulary settings.

In order to solve these challenges, we propose to tackle the task of Open-Vocabulary SGG in a reference-free manner, similar to what has been done in the task of Image Captioning \cite{hesselCLIPScoreReferencefreeEvaluation2021}. By taking advantage of Vision-Language Models such as CLIP \cite{radfordLearningTransferableVisual2021a}, we propose to evaluate the alignment between relation triplets and corresponding image regions formed by the union of subject and object.

\textbf{Weakly Annotated Data.} In contrast to other tasks, such as Object Detection \cite{gupta2019lvis,shao2019objects365}, the task of SGG has not yet seen the emergence of large-scale, high-quality datasets with a large vocabulary. As a result, the main approaches in OV-SGG are using image caption supervision for pre-training \cite{salzmannSceneGraphViTEndtoEnd2025,heOpenVocabularySceneGraph2022a} and then evaluating zero-shot transfer on closed-source datasets such as Visual Genome \cite{krishna2017visual} or PSG \cite{yangPanopticSceneGraph2022}. However, captions-only supervision can limit the understanding of SGG models of fine-grained relations. Several recent approaches have been proposed to alleviate this bias: (i) the RLIPv2 \cite{yuanRLIPv2FastScaling2023} was proposed to generate weakly annotated data by grounding captions to image regions using a dedicated R-TAGGER model; and (ii) the All-Seeing Project V2 (ASv2) \cite{wangAllSeeingProjectV22025} proposed to use Vision-Language Models (i.e. GPT4) to generate open-vocabulary objects and relations annotations on new images. 

Specifically RLIPv2 \cite{yuanRLIPv2FastScaling2023} proposed to generate pseudo-annotations for relations by matching visual regions to predicates in image captions. This strategy comes with a few drawbacks: (i) captions are biased toward certain types of relations (e.g. human actions); (ii) captions are limited in number of relations and typically only contain information about a subpart of the image (e.g. foreground/background); and (iii) the association of relations and image region can fail in images where a lot of similar $<subject, object>$ pairs are present. We detail these biases later in \Cref{sec:evaluation}.

In ASv2 \cite{yuanRLIPv2FastScaling2023}, GPT4 was asked to generate object coordinates, pairs, and predicates in new images on a single prompt. In this strategy, the VLM determines how many and which object pairs to annotate on the image. This can lead to biases since the VLM will be prone to select easy pairs, leaving out rare or difficult relations (e.g. relations between distant objects in the foreground/background). The present paper proposes to use region-specific prompting to ensure global and unbiased coverage of the image content, leading to a more diverse and larger dataset, which can benefit the pre-training of OV-SGG models. We also propose to benchmark the performance of open-source models, in addition to closed-source models such as GPT4.
\section{New Metrics for Open-Vocabulary SGG}
\label{sec:new-metrics}

Evaluating the relevance of a relation to a particular image region is related to Image-Text Alignment and the task of Image-Text Retrieval. The most popular approach to Image-Text Retrieval is the CLIP model\cite{radfordLearningTransferableVisual2021a}. CLIP is a trained image-text pair encoder to align embeddings in a shared space, enabling zero-shot classification and broad generalization across visual and textual tasks. This paper uses CLIP to measure the quality of generated relations.

\subsection{The RelCLIPScore}

Inspired by the CLIPScore metric \cite{hesselCLIPScoreReferencefreeEvaluation2021}, this paper proposes a new reference-free metric, the RelationCLIPScore metric (abbreviated RelCLIPScore) for the Open-Vocabulary SGG task. The CLIPScore metric evaluates the semantic similarity between a generated text (e.g. a caption) and a reference image by using the shared embedding space of the CLIP model \cite{radfordLearningTransferableVisual2021a}, which aligns visual and textual representations. We extend this definition to the prediction of relationships by evaluating the similarity of a predicted $<subject, predicate, object>$ triplet and the corresponding image region defined by the union of the $<subject, object>$ bounding boxes. 
For an image $I$, first the set of predicted union regions that match groundtruth bounding boxes annotations are extracted. Then, for each matching region, the cosine similarity is calculated between CLIP visual embedding and CLIP textual embeddings of the triplet:
\begin{equation}
    \text{CLIPScore}(c,t) = \max(\text{cos}(c,t),0)
\end{equation}
where $c$ is the visual embedding and $t$ the textual embedding. Then, the RelCLIPScore is averaged at the image level:
\begin{equation}
    \text{RelCLIPScore}(I) = \frac{1}{k} \sum_{n=1}^{k} \text{CLIPScore}(c_n, t_n)
\end{equation}
\noindent where $k$ is the number of relations for the image $I$.
Note that a model could attain a high RelCLIPScore by only predicting one relation per image, which is not desirable. To solve this problem, the RelCLIPScore adds a penalty based on the number of predicted relations compared to the maximal number of possible relations to predict given annotated groundtruth objects. Thus, the final RelCLIPScore is  calculated as follows:
\begin{equation}
    \text{RelCLIPScore}(I) = \frac{\frac{1}{k} \sum_{n=1}^{k} \text{CLIPScore}(c_n, t_n)}{\log(p - k)+\alpha} 
\end{equation}

\noindent where $p$ the number of possible relations in the image and $\alpha = 1e^{-5}$.

Note also that, given a set vocabulary of predicates, it is unlikely that every object in the image has at least one relation to every other. Thus, this paper proposes to refine $p$ as follows:
\begin{equation}
    p = \frac{(m * (m-1))}{2}
\end{equation}

\noindent where $m$ is the number of objects.

If references (i.e. ground truth relations) are provided, the Reference RelCLIPScore (Ref-RelCLIPScore) can be computed as follows \cite{hesselCLIPScoreReferencefreeEvaluation2021}:

\begin{multline}
    \text{Ref-RelCLIPScore}(I) = \\
    \frac{2 * \text{CLIPScore}(c_n, t_n) *  \text{SimText}(t_{gt}, t_{pred})}{ \text{CLIPScore}(c_n, t_n) *  \text{SimText}(t_{gt}, t_{pred})}
\end{multline}

\noindent with 
\begin{equation}
    \text{SimText}(t_1, t_2) = \max(\text{cos}(t_{1},t_{2}),0)
\end{equation}

\noindent where $t_{gt}, t_{pred}$ are textual embeddings of the target and prediction, respectively. Note that differences between the original CLIPScore \cite{hesselCLIPScoreReferencefreeEvaluation2021} and the proposed RelCLIPScore are introduced with equations (2), (3), and (4).



Recently several new approaches to Image-Text Retrieval have been proposed such as BLIP-2 \cite{liBLIP2BootstrappingLanguageImage2023} and SIGLIP \cite{zhaiSigmoidLossLanguage2023}. Similar to CLIP, these approaches can be used for image-text alignment. 
In addition to image-text distances, BLIP-2 computes \textit{probabilities} as an Image-Text Matching (ITM) score to learn fine-grained alignment between textual and visual representation.  Yuksekgonul et. al. \cite{yuksekgonuland} proposed NegCLIP, a revised version of CLIP trained with hard negatives for better image-text alignment in challenging scenarios. 

In order to complement the RelCIPScore, this paper proposes also the RelBLIPScore and RelSIGLIPScore to benchmark relations generation. These scores are not computed using cosine similarity but rather by taking the ITM score of BLIP or the probabilities computed by the Sigmoid function for SIGLIP.

In the following, we present an evaluation of these different metrics and their respective alignment with human annotations in leading datasets.



\subsection{Alignment with Ground truth Annotations}

In order to evaluate the usage of the new RelCLIPScore metric proposed in this paper, we computed its alignment with groundtruth annotations in the PSG \cite{yangPanopticSceneGraph2022} and VG \cite{krishna2017visual} datasets. For each groundtruth relation, we computed the CLIPScore (or BLIPScore / SIGLIPScore) of the triplet as well as the score of any other possible triplet with the same $<subject, object>$ pair but with a different predicate. In this case, we only took predicates that are annotated at least once in the dataset for the corresponding $<subject, object>$ pair. In order to take into account the ambiguity of predicate annotations \cite{tangUnbiasedSceneGraph2020b,neauFineGrainedTooCoarse2023}, we used the rank of the groundtruth predicate score to compute a matching score $\theta$:
\begin{equation}
    \theta(p) =  1 - \frac{\mathrm{rank}^{(0)}_L(p)}{|L|}
\end{equation}

\noindent where $p$ is the score of the groundtruth predicate and $L$ the list of scores of all possible predicates for the given pair.

Note that, if the score of the groundtruth predicate is close to the highest score but not the highest because of a similar predicate (for instance with the confused pair \textit{on / on top of}), then its matching score $\theta$ will still be high.

\begin{table}[h!]
    \caption{Comparison of the different image-text alignment metrics for visual relationship prediction on the PSG \cite{yangPanopticSceneGraph2022} and VG \cite{krishna2017visual} datasets.}
    \label{tab:metrics_comparison}
    \resizebox{\columnwidth}{!}{
    \renewcommand{\arraystretch}{1.1}
        \centering
        \begin{tabular}{c|c|c|c|c|c}
            \hline 
            \textbf{D} & \textbf{Model} & \textbf{Method} & \textbf{Score} & \textbf{Prec.} & \textbf{$\theta$} \\
            \hline
            \multirow{4}{*}{\rotatebox{90}{\textbf{PSG}}} & 
            NegCLIP \textsubscript{B-32} \cite{yuksekgonuland} & Cosine & 24.47 & \textbf{30.18} & \textbf{66.10} \\
            & CLIP \textsubscript{L-14} \cite{radfordLearningTransferableVisual2021a} & Cosine & 23.64 & 27.77 & 63.14 \\
            & SIGLIP \cite{zhaiSigmoidLossLanguage2023} & Sigmoid & 17.78 & 25.23 & 62.36 \\
            & BLIP2 \cite{liBLIP2BootstrappingLanguageImage2023} & ITM & 67.38 & 24.19 & 62.29 \\
            \hline
            \hline
            \multirow{4}{*}{\rotatebox{90}{\textbf{VG}}} & 
            NegCLIP \textsubscript{B-32} \cite{yuksekgonuland} & Cosine & 18.84 & 22.75 & 64.98 \\
            & CLIP \textsubscript{L-14} \cite{radfordLearningTransferableVisual2021a} & Cosine & 20.73 & 23.80 & 61.85 \\
            & SIGLIP \cite{zhaiSigmoidLossLanguage2023} & Sigmoid & 23.69 & 17.55 & 58.66 \\
            & BLIP2 \cite{liBLIP2BootstrappingLanguageImage2023} & ITM & 65.20 & 9.38 & 53.49 \\
            \hline
        \end{tabular}
    }

\end{table}

\cref{tab:metrics_comparison} compares the different models and their alignment with ground truth data, reporting precision as the proportion of groundtruth predicates with top score over all samples. 

Note that this metric is less representative than the matching score ($\theta$) because of confused predicates.
We observed that the NegCLIP model is the most performant for image region-visual relationship alignment with a matching score ($\theta$) of 66.10 on the PSG dataset and 64.98 on VG. Despite being more recent, approaches with dedicated image-text alignment scores (i.e. sigmoid and ITM), SIGLIP and BLIP2 are worse at ranking candidate relations for a specific image region. 

\begin{table}[t]
    \caption{Top confused triplets in the PSG dataset.}
    \label{tab:confused_triplets}
    \centering
    \resizebox{\columnwidth}{!}{
    \begin{tabular}{c|c}
        \hline
        \textbf{Groundtruth Triplet} & \textbf{Confused triplet} \\
        \hline
        \textit{person in front of person} & \textit{person beside person} \\
        \textit{person talking to person} & \textit{person beside person} \\
        \textit{person looking at person} & \textit{person beside person} \\
        \textit{person holding baseball bat} &  \textit{person swinging baseball bat} \\
        \textit{book on shelf} & \textit{book in shelf} \\
        \hline
    \end{tabular}
    }

\end{table}

\Cref{tab:confused_triplets} shows the top 5 confused triplets in the PSG dataset from the CLIP model (the confused triplets have a higher Score than the groundtruth). 
Note that, due to the fact that predicates are not exclusive in the dataset, the confused triplets are likely not false positives but ambiguous true positives (for instance \textit{person talking to person} and \textit{person beside person} are likely both valid relations).

\subsection{Limitations}

After further analysis, we identified different failing cases when evaluating the performance of the CLIP, SIGLIP, and BLIP models for image-relations alignment. 

\begin{figure}[t!]
    \centering
    \begin{subfigure}[t]{.5\linewidth}
            \includegraphics[width=\textwidth]{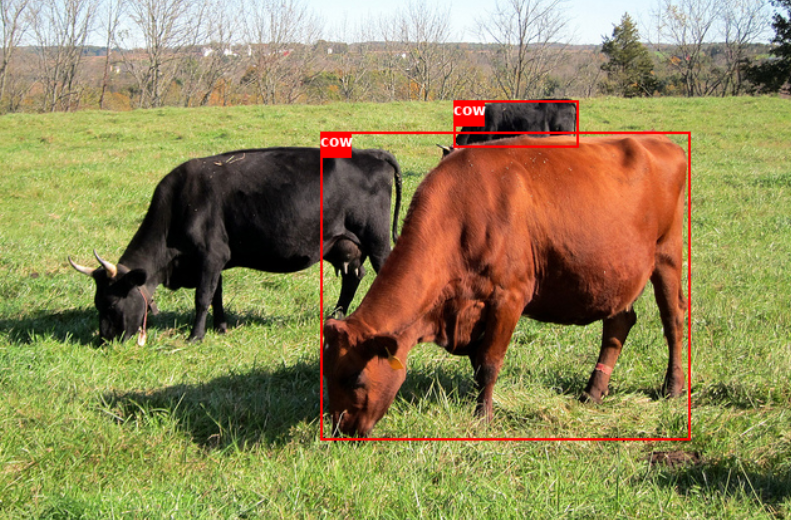}\label{fig:img_cows}
            \caption{}
    \end{subfigure}
    \hfill
    \begin{subfigure}[c]{\linewidth}
        \includegraphics[width=\textwidth]{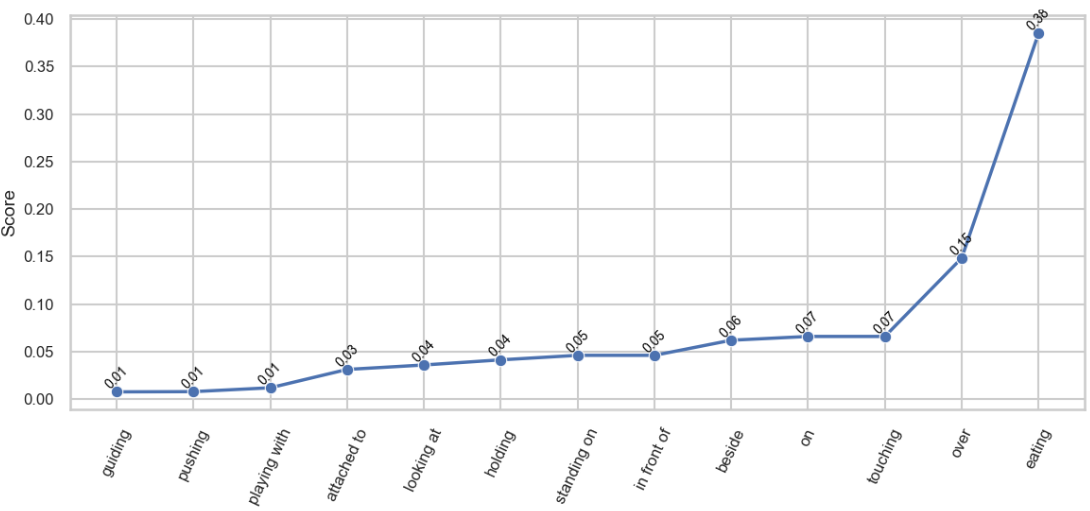}\label{fig:img_score_ranking}
        \caption{}
    \end{subfigure}
    \hfill
    \caption{(a) is the original image and (b) the CLIP ranking of all possible predicates for the pair $<$\textit{cow\_1, cow\_2}$>$.}
    \label{fig:cows_example}
\end{figure}

First, these models can struggle to identify the correct predicate when the $subject$ or the $object$ of the relation is extensively bigger in the image than its pair. \Cref{fig:cows_example} shows an example: the image contains two cows, but the first one is significantly bigger than the second, so the CLIP model is asked to assign a confidence score for each possible predicates between the two animals, it generates that top score is $<cow, eating, cow>$ with 0.38, where the second highest $<cow, over, cow>$ (which is more realistic) is only given a score of 0.16. One could hypothesize that the model might be biased by the action that the first cow is doing (i.e. $eating$) and overlooking the relation with the other cow. 

\begin{figure}[t!]
    \centering
    \begin{subfigure}[t]{.5\linewidth}
            \includegraphics[width=\textwidth]{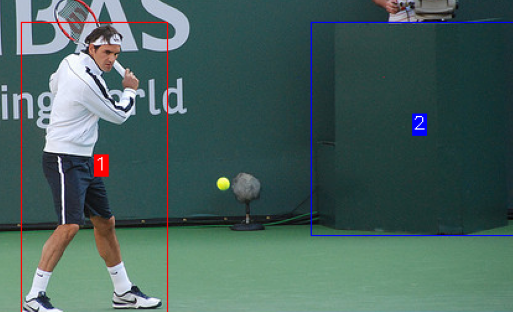}\label{fig:img_distant}
            \caption{}
    \end{subfigure}
    \hfill
    \begin{subfigure}[c]{\linewidth}
        \includegraphics[width=\textwidth]{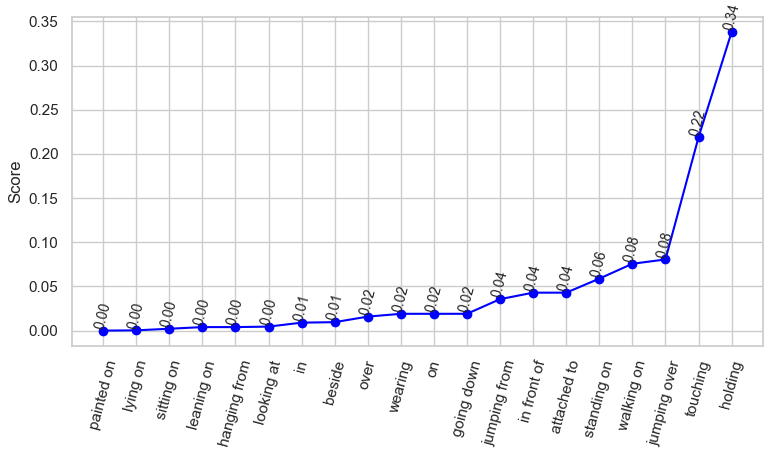}\label{fig:img_distant_scores}
        \caption{}
    \end{subfigure}
    \hfill
    \caption{(a) is the original image and (b) the CLIP ranking of all possible predicates for the pair $<$\textit{person\_1, wall\_2}$>$.}
    \label{fig:distant_example}
\end{figure}

In our analysis, we also observed similar difficulties when the models intend to rank predicates when the subject and object are two distant objects in the image. \Cref{fig:distant_example} displays an example with the pair $<person, wall>$ where the correct predicate is \textit{in front of}, but the top score predicates are \textit{holding}, \textit{touching}, or \textit{jumping over} which not describing the real image content. This shows the difficulty of the model to focus on the correct regions of the image when there is a significant distance between the subject and object of the relation.

\section{Evaluating Vision-Language Models for Region-Specific Relationship Prediction}
\label{sec:evaluation}

In this section, the introduced RelCLIPScore metric is used to evaluate the performance of a selection of VLMs on the task of relationship prediction on image regions. VLMs are traditionally evaluated at the image level on downstream tasks such as Image Captioning \cite{yangTransformingVisualScene2023}, VQA \cite{schaferVisualAnalysisSceneGraphBased2023}, or Visual Reasoning \cite{thrush2022winoground}. In these tasks, the evaluation of relations is implicit, either contained in captions or questions. The performance of Vision-Language Models for relation prediction at the triplet level (i.e. given only the $<subject, object>$ pair and associated image region) remains unknown. 


As a result, we present an analysis of the performance of VLMs for relationship predictions on regions of interest with subject-object prompting. By prompting VLMs per region of images, we can measure their limitations in challenging scenarios, for instance with small or distant objects.

\subsection{Method}

\textbf{Models.} We selected a set of Open-Source and closed-source models for our comparison. For Open Source models, we used the LlaVa-OneVision 7B variant \cite{li2024llava}, InterVL3-8B \cite{zhu2025internvl3} and Qwen-2.5VL 7B \cite{baiQwen25VLTechnicalReport2025}. We compared these models against the closed-source GPT4o-mini and GPT4o from OpenAI \footnote{\url{https://openai.com/index/gpt-4o-system-card/}}.

\textbf{Datasets.}  The PSG dataset was used \cite{yangPanopticSceneGraph2022} to evaluate the different models, which contains 46K images in the training set and 2K images in the test set. These images were selected from the COCO dataset and carefully annotated by humans with bounding boxes and relations. This dataset is deemed of better quality than other SGG datasets such as Visual Genome or GQA \cite{yangPanopticSceneGraph2022}. We evaluated each model in zero-shot settings on the test set of PSG. 

\textbf{Relation Prediction.} For relation prediction, we employed the Set-of-Mark prompting strategy \cite{yangSetofMarkPromptingUnleashes2023} by annotating the subject and object masks on the image before feeding it to the model. We used the Segment-Anything Model (SAM)  \cite{kirillov2023segment} to generate masks, with a consistent color for subject (blue) and object (red).  
In addition, images are cropped on the union of the subject-object boxes, extended by 20\% to provide context to the model. The subject of the relation is referred to as ``Object 1" in the prompt, and the object of the relation is ``Object 2".

\subsection{Results}

We measured the CLIPScore, RefCLIPScore and overall precision on the PSG dataset. \Cref{tab:models_comparison} provides the results obtained. 
Note that the LlaVa-OneVision model is the best by a small margin with respect to the NegCLIP score, outperforming even closed-source models such as GTP4o. Note also that there is a high variation in precision for all models. However, other metrics (reference-free or not) show less variations, which indicate that precision is not a good indicator to evaluate VLMs in Open-Vocabulary SGG. Some models, such as InternVL, are more aligned with ground-truth annotations (see Precision column in \Cref{tab:models_comparison}), whereas others (LlaVa, GPT4o-mini) are more aligned with the visual content, if we refer to the NegCLIP, CLIP, and SIGLIP scores, for instance. If we compare these results with scores computed on groundtruth annotations (see \Cref{tab:metrics_comparison}), we can see that the best model in NegCLIP score (24.02) is still behind the NegCLIP score of groundtruth (24.47), and similarly for other metrics, which shows the current limitations of VLMs. In the following, we analyse further  these limitations in a series of ablation studies.

\begin{table*}[!ht]
    \caption{Comparison of the different image-text alignment metrics for Visual Relationships prediction on the PSG dataset \cite{yangPanopticSceneGraph2022}.}
    \label{tab:models_comparison}
    \centering
    \renewcommand{\arraystretch}{1.1}
    \begin{tabular}{c|c|c|c|c|c}
        \hline 
        \multirow{2}{*}{\textbf{Model}} & \multicolumn{4}{c|}{\textbf{Score}} \\
        & \textbf{NegCLIP} & \textbf{RefNegCLIP} & \textbf{CLIP} & \textbf{SIGLIP} & \textbf{Precision} \\
        \hline
        LlaVa-OneVision 7B \cite{li2024llava} & 24.02 & 47.75 & 19.52 & 10.02 & 16.07  \\
        InternVL-3 8B \cite{zhu2025internvl3} & 23.94 & 47.73 & 19.60 & 11.26 & 30.97 \\
        Gwen2.5VL 7B \cite{baiQwen25VLTechnicalReport2025} & 23.88 & 47.98 & 19.55 & 9.90 & 14.01 \\
        \hline
        \hline
        GPT4o-mini & 24.00 & 47.88 & 19.81 & 11.27 & 16.23 \\
        GPT4o & 23.97 & 47.76 & 19.82 & 10.38 & 18.05 \\
        \hline
    \end{tabular}

\end{table*}

\subsection{Ablation Studies}
\begin{figure}[t]
    \centering
    \includegraphics[width=\linewidth]{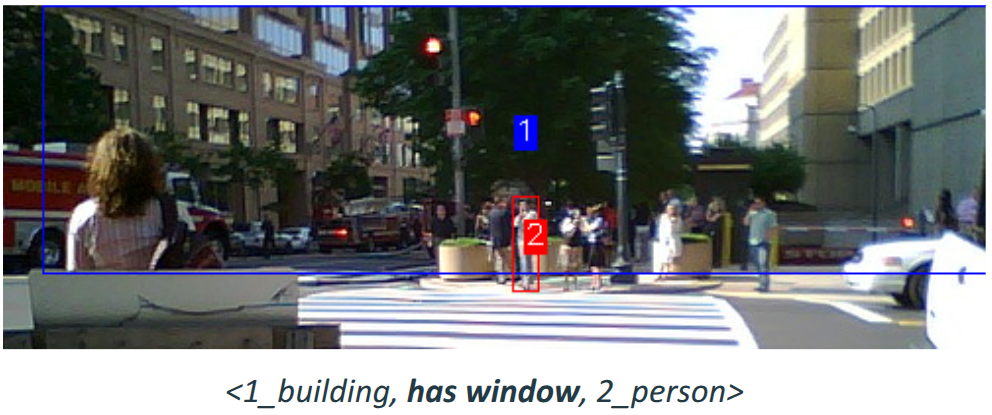}
    \caption{Example of the pair imbalance problem. Relation generated using the LlaVa-OV-7B model \cite{li2024llava}.}
    \label{fig:pair_imbalance}
\end{figure}

\begin{table}[t]
    \caption{Box ratio experiments, Low Ratio corresponds to $<1:5$ ratio between subject and object whereas High Ratio corresponds to $>1:5$.}
    \label{tab:ratio_expe}
    \centering
    \resizebox{\columnwidth}{!}{
        \renewcommand{\arraystretch}{1.1}
        \begin{tabular}{c|c|c|c|c}
        \hline
             \multirow{2}{*}{\textbf{Model}} & \multirow{2}{*}{\textbf{Ratio}} & \multicolumn{2}{c|}{\textbf{Score}} & \multirow{2}{*}{\textbf{Prec.}} \\
            & & NegCLIP & RefNegCLIP & \\ \hline
            \multirow{2}{*}{InternVL3-8B} & Low & 23.91 & 47.45 & 29.22 \\ 
            ~ & High & 24.15 & 47.97 & 40.86 \\ \hline
            \multirow{2}{*}{LlaVa-OV-7B} & Low & 23,89 & 47.46 & 10,65 \\ 
            ~ & High & 24,11 & 48.01 & 23.88 \\ \hline
        \end{tabular}
    }
\end{table}
\textbf{Pair size imbalance.} As we have seen in the previous section, Vision-Language Models can easily struggle to find the correct relation when there is a significant imbalance between the respective sizes of the subject and object in the image. To quantify this limitation, we ran a new experiment by selecting a subset of 1,000 pairs in the PSG dataset where the boxes ratio is $>1:5$. We compared performance on this subset with a second subset of 1,000 pairs, where the boxes ratio is $<1:5$. \Cref{tab:ratio_expe} summarizes the results of our experiments. Note that there is  a consistent decline in RelSCORE and Precision metrics for pairs with size imbalance (low ratio settings) compared to pairs of similar size (high ratio settings). These findings highlight the struggle of VLMs to perceive smaller objects with good accuracy \cite{yuksekgonuland}. \Cref{fig:pair_imbalance} provides an example of a relation generated with pair imbalance (boxes ratio $=0.04$). Note that the VLM is not able to focus on the object of the relation (\textit{2\_person}) and instead only describes an attribute of the subject (\textit{1\_building}).

\textbf{Distant objects.} SGG benchmarks \cite{krishna2017visual,yangPanopticSceneGraph2022} contain mostly annotations of object pairs that intersect. However, in real-world applications \cite{schaferVisualAnalysisSceneGraphBased2023,amodeoOGSGGOntologyGuidedScene2022}, describing relations between distant objects can be critical. Generating relations between distant objects can also be important to augment the robustness of the pre-training in Open-Vocabulary SGG.
In these experiments, we extracted a subset of 1,000 pairs of objects separated by at least 20\% of the image size and compared the performance of VLMs with another subset of 1,000 pairs that intersect (i.e. $IoU > 0$).
\Cref{tab:distant_expe} shows the results of these experiments. Note that there is a significant drop in all metrics for pairs that do not intersect compared with pairs that do.

\begin{figure}
    \centering
    \begin{subfigure}[t]{.6\linewidth}
            \includegraphics[width=\textwidth]{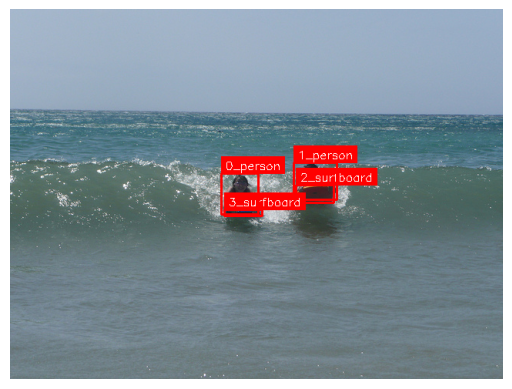}\label{fig:img_surf}
    \end{subfigure}
    \hfill
    \begin{subfigure}[c]{.4\linewidth}
        \includegraphics[width=\textwidth]{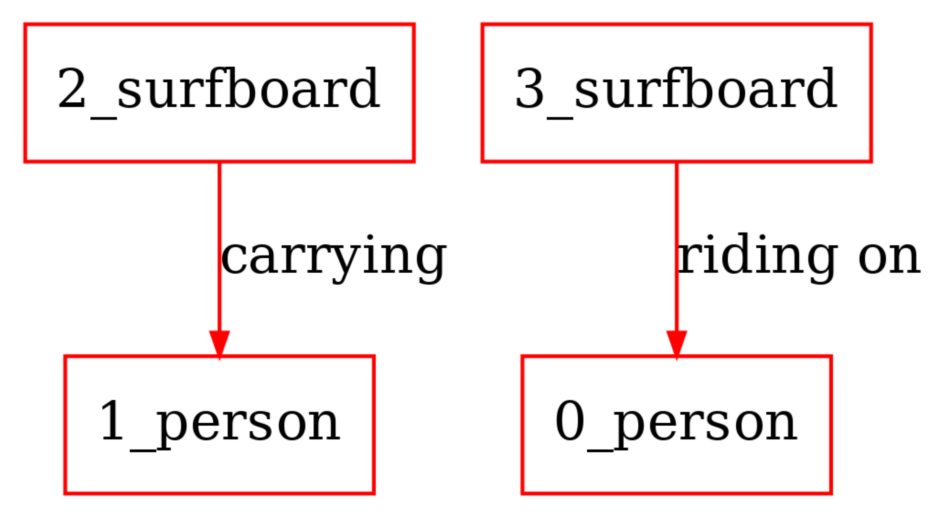}\label{fig:surf_llava}
        \caption{Predictions from LlaVa.}
    \end{subfigure}
    \hfill
    \begin{subfigure}[c]{.5\linewidth}
        \centering
        \includegraphics[width=\textwidth]{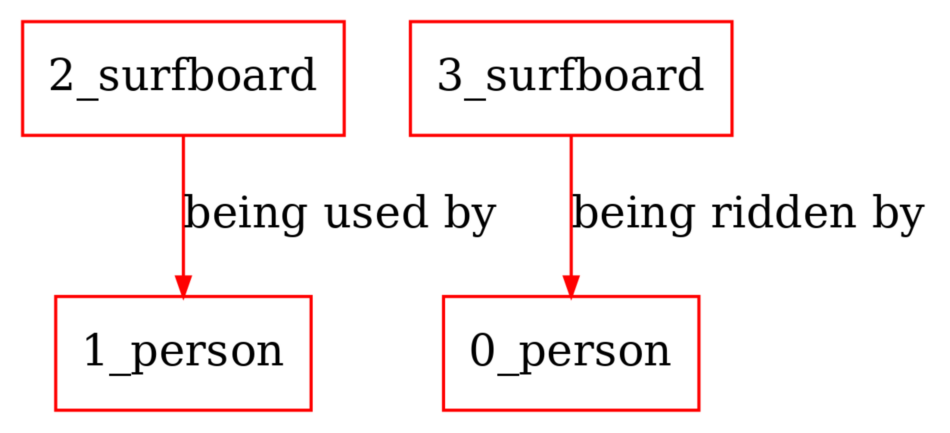}\label{fig:surf_gt}
        \caption{Groundtruth.}
    \end{subfigure}
    \hfill
    \caption{The inverse relationship bias: some pairs are used to be seen in mono-directional relations. VLMs can then struggle to pay attention to the directionality of the relation.}
    \label{fig:example_surf}
\end{figure}

\textbf{Inverse Relations.} We also observed that for certain types of relations (mainly human actions), VLMs can be confused by the directionality of the relation. \Cref{fig:example_surf} provides an example of this issue: the pair $<surfboard, person>$ is prompted in a direction which is unusual (i.e. $<person, riding, surfboard>$ is more likely to appear in the training data of VLMs than $<surfboard, being \ ridden \ by, person>$), resulting in a failure from the LlaVa model. However, quantifying this issue is complex because not all $<human, object>$ pairs are biased toward one direction. However, even with these shortcomings, we believe that VLMs can be useful for generating pre-training data for the task of Open-Vocabulary SGG. In the next section, we evaluate this hypothesis with the introduction of the FG-OV dataset.

\begin{table}[t]
    \caption{Experiments with distant objects: Intersect corresponds to IoU $>= 0$ and Non-Intersect to IoU $< 0$.}
    \label{tab:distant_expe}
    \centering
    \resizebox{\columnwidth}{!}{
        \renewcommand{\arraystretch}{1.1}
        \begin{tabular}{c|c|c|c|c}
        \hline
             \multirow{2}{*}{\textbf{Model}} & \multirow{2}{*}{\textbf{Intersect}} & \multicolumn{2}{c|}{\textbf{Score}} & \multirow{2}{*}{\textbf{Prec.}} \\
            & & NegCLIP & RefNegCLIP & \\ \hline
            \multirow{2}{*}{InternVL3-8B} & Yes & 23.78 & 47.27 & 36.13 \\ 
            ~ & No & 23.01 & 45.53 & 30.08 \\ \hline
            \multirow{2}{*}{LlaVa-OV-7B} & Yes & 24.01 & 47.34 & 11.67 \\ 
            ~ & No & 22.98 & 45.65 & 3.7 \\ \hline
        \end{tabular}
    }
\end{table}



\section{The FG-OV SG Dataset}
\label{sec:FG-OV}

This section presents our new method for generating synthetic data for the Open-Vocabulary SGG task. Inspired by previous work \cite{yuanRLIPv2FastScaling2023}, we used images from the COCO and Objects365 datasets. The COCO dataset \cite{lin2014microsoft} contains 122,218 images annotated with bounding boxes for 80 object classes. The Objects365 dataset \cite{shao2019objects365} contains more than 1M images annotated with bounding boxes for 365 object classes. Our goal is to generate high-quality synthetic relation annotations between object pairs to create a new SGG dataset.
To do so, we used a combination of all images from COCO and 80K images from the train set of OBJ365, for a total of 200,218 images.

An important concern when creating Scene Graph datasets is the diversity of predicates per $<subject, object>$ pair. In fact, if a pair of objects is mainly associated with a single predicate, then it is easy for a neural network to overfit to this single predicate, hindering generalization capabilities \cite{zellersNeuralMotifsScene2018a}. Empirically, we experienced a high recall for pairs associated with a low number of predicates and a low recall for the opposite. 
In Open-Vocabulary settings, every object can possibly have at least one relation with every other object, leading to a complexity of $m * (m-1)$. For datasets with a large number of object classes, such as COCO or Objects365, annotating all pairs will be too resource-intensive. As we have seen in the previous section, VLMs can struggle to generate relations with distant objects; thus, we selected pairs of objects that intersect (i.e. $IoU > 0$) as valid candidates for our approach. Finally, we randomly selected 50\% of those pairs for relation prediction with a maximum of 50 per image. For the COCO dataset, this process results in an average of 9 relations generated per image; for the Objects365, the average is 17.

We used the LlaVa-OneVision 7B model \cite{li2024llava} to generate the data, using the region prompting strategy introduced in \Cref{sec:evaluation}. We post-processed generations by removing predicates with a length greater than 5 words. In the prompt, we also indicated that generating vague predicates (e.g. $next \ to$) was not desirable. In the next sections, we compare this approach with previous work.

\subsection{Data quality}

In comparison with RLIPv2 \cite{yuanRLIPv2FastScaling2023}, both data mixtures are very similar in size; the main difference is the quality and diversity of relations, as shown in \Cref{tab:data_mixture}. Note that our split generated on the COCO dataset contains 3x more predicates. The high number of predicates in the VG dataset comes from the fact that the raw data is used, containing a high number of noisy annotations \cite{zhang2019large} (e.g. \textit{$<$person, has a small light blue handbag attached to, hand$>$} is a triplet in VG).

\begin{table}[t!]
    \caption{Data mixture comparison.}
    \label{tab:data_mixture}
    \centering
    \resizebox{\columnwidth}{!}{
    \renewcommand{\arraystretch}{1.1}
        \begin{tabular}{c|c|c|c|c}
            \textbf{Mixture} & \textbf{Dataset} & \textbf{Images} & \textbf{Triplets} & \textbf{Predicates} \\
            \hline
            \multirow{2}{*}{RLIPv2} & VG & 108,077 & 1,987,331 & 36,515 \\
            & COCO & 122,218 & 902,278 & 388 \\
            \hline
            \multirow{2}{*}{Ours} & VG & 108,077 & 1,987,331 & 36,515 \\
            & COCO & 122,218 & 819,111 & 1,121 \\
        \end{tabular}
    }

\end{table}

\begin{figure}[t!]
    \centering
    \includegraphics[width=\columnwidth]{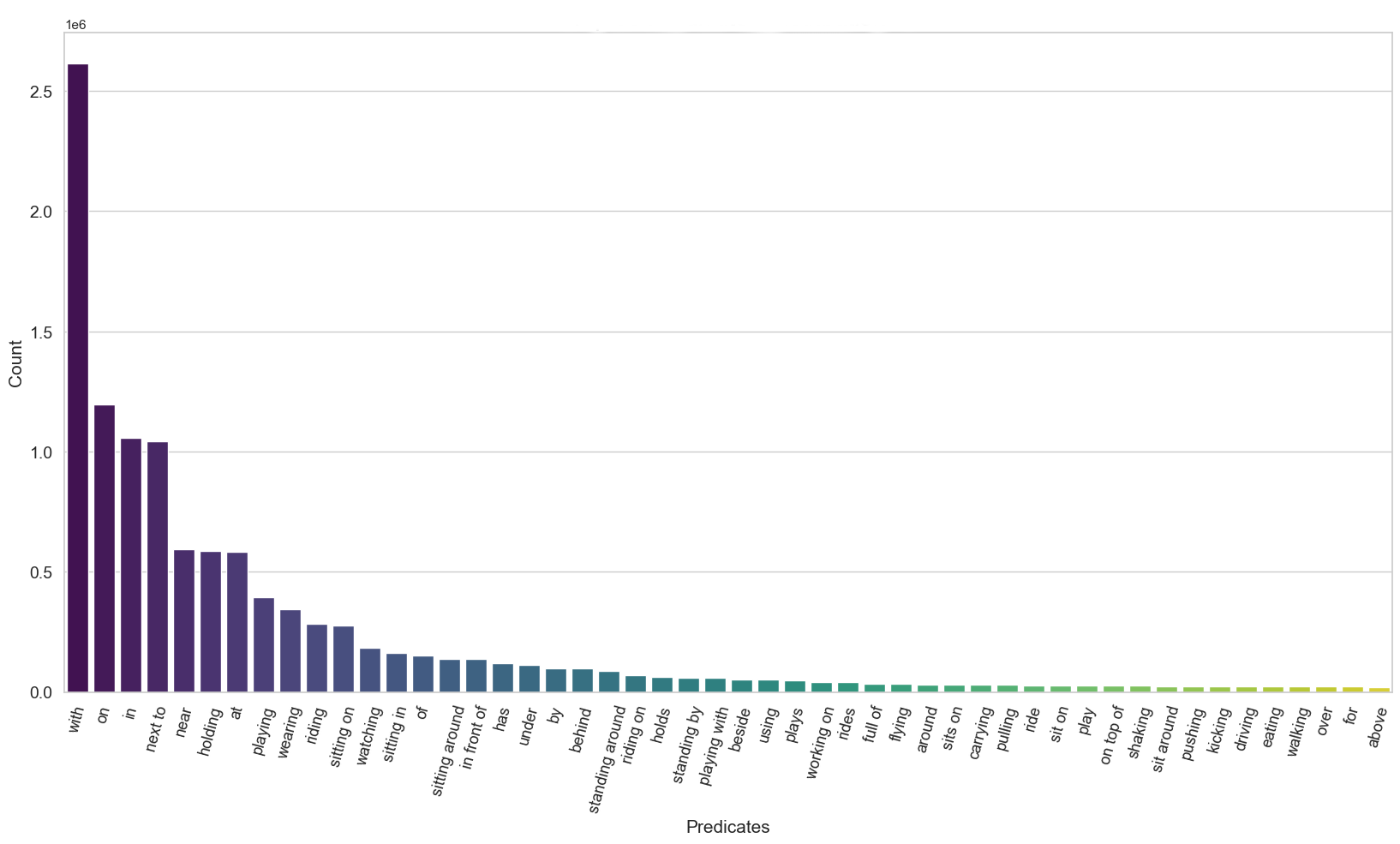}
    \caption{Predicate distribution of RLIPv2 (OBJ365 images), we can observe that the top-5 predicates are not fine-grained.}
    \label{fig:distri_rlipv2}
\end{figure}

\begin{figure}[t!]
    \centering
    \includegraphics[width=\columnwidth]{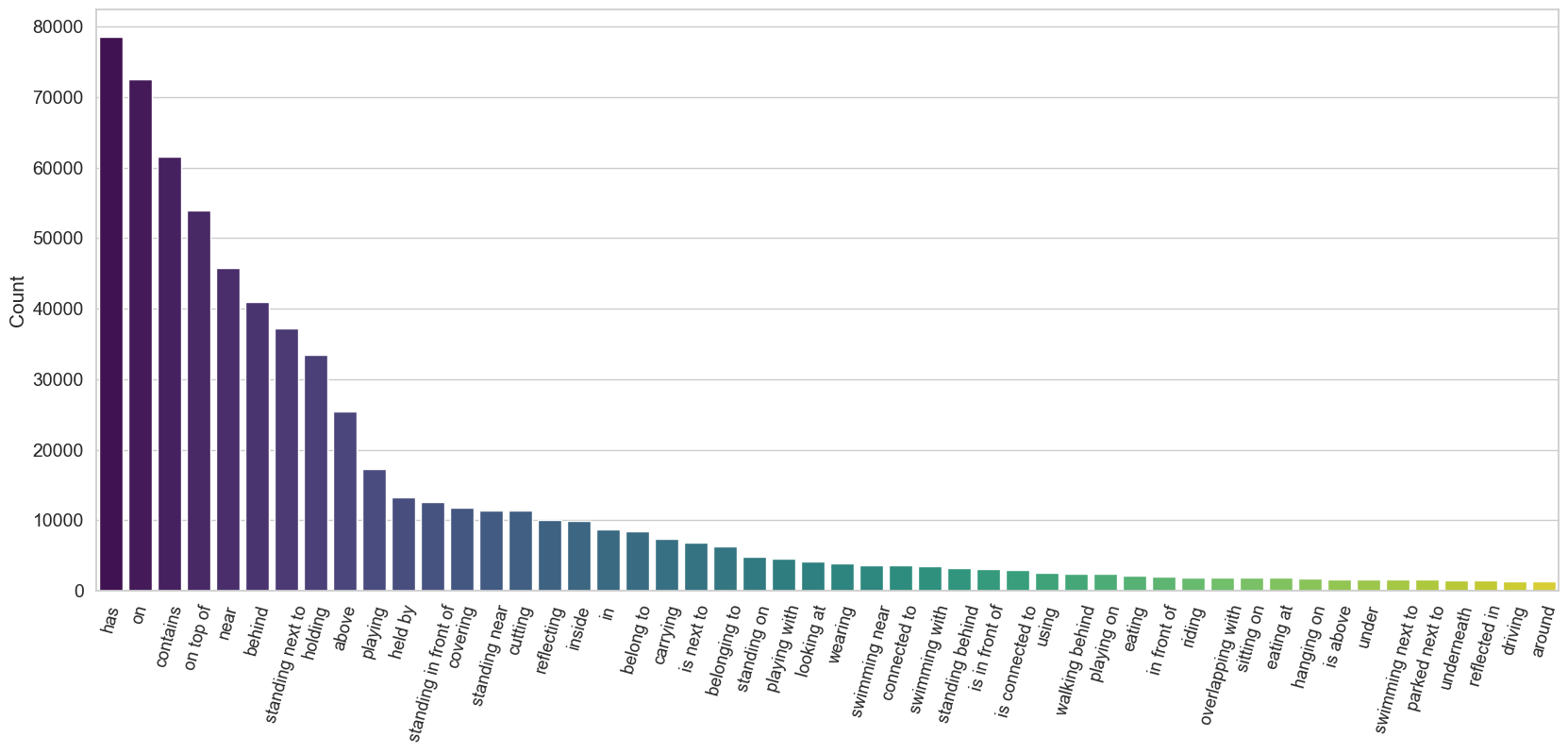}
    \caption{Predicate distribution of the FG-OV dataset (100K images from OBJ365). Compared to RLIPv2, we can observe a better balance of the distribution as well as more fine-grained predicates.}
    \label{fig:distri_ours}
\end{figure}

\Cref{fig:distri_rlipv2} and \Cref{fig:distri_ours} compare the predicate distribution of both data splits. One important concern with SGG is the long-tail distribution of predicates, which can significantly hurt the training of neural networks \cite{tangUnbiasedSceneGraph2020b}. \Cref{fig:distri_ours} shows that the predicates in our data split are more diverse and balanced than those from RLIPv2 (see \Cref{fig:distri_rlipv2}). For RLIPv2, note that the top predicates are not fine-grained (e.g. ``with", ``on", ``next to"). Pre-training Open-Vocabulary SGG models on coarse data may hurt the generalization of such models \cite{tangUnbiasedSceneGraph2020b}. 

\Cref{tab:rlipv2_results} presents the quality of annotations generated using our method in contrsat to the data generated in RLIPv2 \cite{yuanRLIPv2FastScaling2023}. 
We observed higher NegCLIP, CLIP, and SIGLIP scores with our data split, when evaluated on the same set of images. These results demonstrate better region-triplet alignment in the data generated using our region prompting strategy.

\begin{table}[t!]

    \caption{Comparison of the region-triplet alignment of data used in RLIPv2 and our generation with the LlaVa model. Scores are computed on the test set of the COCO dataset.}
    \label{tab:rlipv2_results}
    \centering
    \renewcommand{\arraystretch}{1.1}
    \resizebox{.8\columnwidth}{!}{
    \begin{tabular}{c|c|c|c}
        \hline 
        \multirow{2}{*}{\textbf{Data Split}} & \multicolumn{3}{c}{\textbf{Score}} \\
        & \textbf{NegCLIP} & \textbf{CLIP} & \textbf{SIGLIP} \\
        \hline
        RLIPv2 \cite{yuanRLIPv2FastScaling2023} & 25.41 & 22.22 & 18.50 \\
        FG-OV (ours) & 25.97 & 22.81 & 20.72 \\        
        \hline
    \end{tabular}
    }

\end{table}

\begin{figure}[t!]
    \centering
    \begin{subfigure}[t]{.6\linewidth}
            \includegraphics[width=\textwidth]{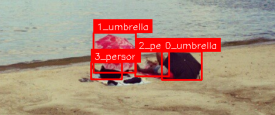}\label{fig:img_source}
            \caption{Original image.}
    \end{subfigure}
    \hfill
    \begin{subfigure}[c]{.5\linewidth}
        \includegraphics[width=\textwidth]{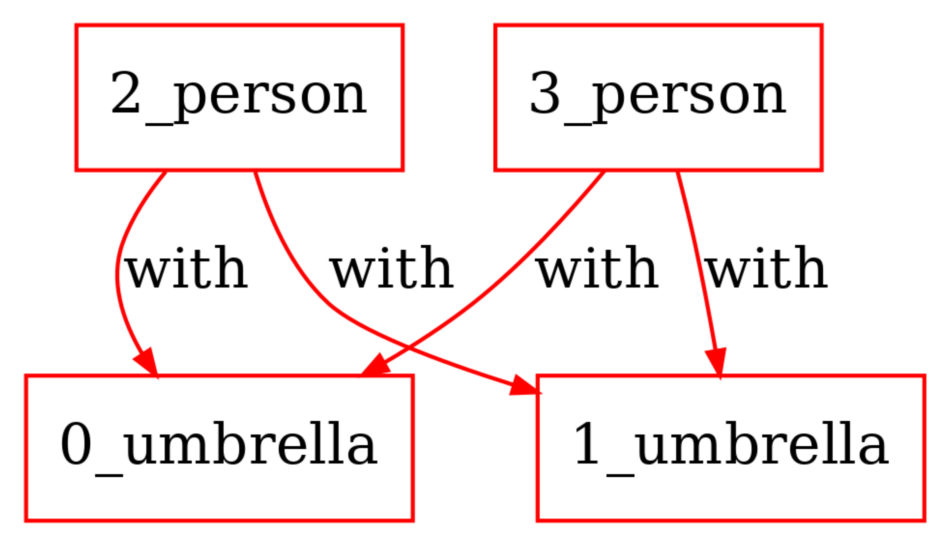}\label{fig:rlip_graph}
        \caption{Data from RLIPv2 \cite{yuanRLIPv2FastScaling2023}.}
    \end{subfigure}
    \hfill
    \begin{subfigure}[c]{.4\linewidth}
        \centering
        \includegraphics[width=\textwidth]{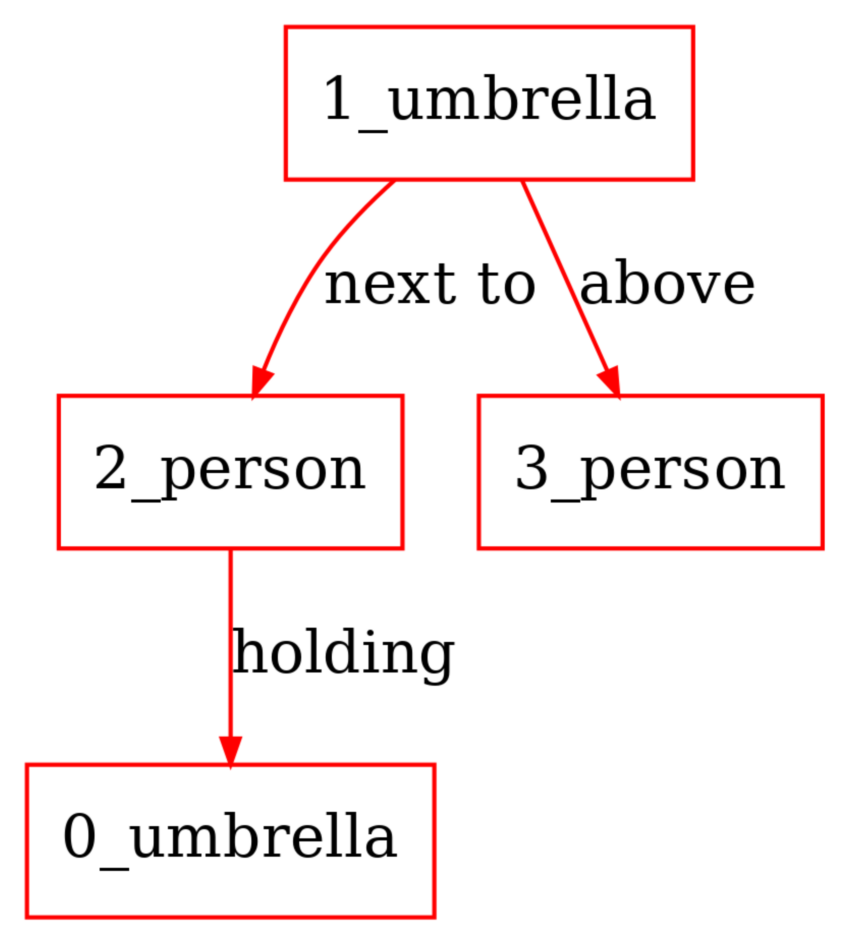}\label{fig:llava_graph}
        \caption{Data from FG-OV (ours).}
    \end{subfigure}
    \hfill
    \caption{Comparison between data from the baseline RLIPv2 and our method using region-specific prompting with LlaVa-OneVision.}
    \label{fig:predictions_sgdiff}
\end{figure}

\Cref{fig:predictions_sgdiff} compares annotations from the RLIPv2 dataset and our split generated using the LlaVa model. Due to ambiguous captions (i.e. \textit{a person with an umbrella on the beach}), the relations from RLIPv2 are coarse and uninformative in this example. Our data split is more fine-grained and contains a precise description of both the topological layout of the scene (e.g. $<umbrella, next \ to, person>$, $<umbrella, above, person>$) and the functional layout (e.g. $<person, holding, umbrella>$). By comparing this example with the distribution of relations in \Cref{fig:distri_rlipv2}, we can observe that spurious annotations (as in this previous example) are present to a significant extent in the data because the $with$ predicate class is overly represented.

\subsection{Open-Vocabulary SGG Results}

In addition to evaluating the data quality, we evaluated the benefit of using the FG-OV dataset for pre-training of OV-SGG models. We replicated the training of the RLIPv2 model in two settings: 
\begin{enumerate}[i.]
    \item Original pre-training \cite{yuanRLIPv2FastScaling2023}, using a data mixture composed of Visual Genome and COCO, where the annotations for COCO were generated by the RLIP method (R-PARSER).
    \item Our pre-training, using the same data mixture but with COCO annotations generated by region-specific prompting with LlaVa (i.e. part of our FG-OV dataset).
\end{enumerate}

We pre-trained the RLIPv2 model for 5 epochs for each data mixture using the codebase and parameters given by the original authors \cite{yuanRLIPv2FastScaling2023}.
Then, we evaluated our pre-training strategy for transfer learning on the HICO-DET dataset \cite{chao2018learning}. Similar to previous work \cite{yuanRLIPv2FastScaling2023}, we report results for zero-shot on the Unknown Combination (UC) settings UC-RF and UC-NF. UC-RF refers to rare combinations, and UC-NF refers to non-rare combinations. \Cref{tab:zero-shot_UC-NF_UC-RF} shows that pre-training the model with our data split consistently increases the performance. The performance of RLIPv2 data mixture is different from the original work \cite{yuanRLIPv2FastScaling2023} because we only pre-trained the model for 5 epochs instead of 20. 
For rare combinations (UC-RF), we see an important gain for the Unseen set (full zero-shot) from 15.88 to 16.99\%. The average relative improvement for all metrics is +3.5\%. This demonstrates the importance of the diversity of pretraining data for Open-Vocabulary SGG.

\begin{table}[t!]
    \caption{Comparisons on HICO-DET under UC-RF and UC-NF settings. Results are reported on \textit{Unseen}/\textit{Seen}/\textit{Full} sets.}
  \label{tab:zero-shot_UC-NF_UC-RF}
  \centering
    \resizebox{\columnwidth}{!}{
        \begin{tabular}{c|c|c}
        \toprule
        \textbf{Data Mixture}  & \textbf{UC-RF} & \textbf{UC-NF} \\
        \midrule
        \midrule
        RLIPv2 & 15.88 / 26.00 / 23.98 &  18.87 / 19.87 / 19.67 \\
        FG-OV (ours) & 16.99 / 25.92 / 24.12 & 18.77 / 21.27 / 20.77 \\
        \bottomrule
        \end{tabular}
    }
\end{table}
\section{Conclusion}
\label{sec:conclusion}

This paper presented threefold contributions to the Open-Vocabulary SGG task. First, we defined a new reference-free metric, the RelCLIPScore, which can be used with different models (NegCLIP, CLIP, or SIGLIP) to evaluate image-text alignment at the relation level. Second, we proposed a review and evaluation of a selection of VLMs on the task of relation prediction from image regions, showing the shortcomings of current state-of-the-art models in challenging scenarios. We especially witnessed limitations for pairs with size imbalance or distant repartition at the image level. Then, we used our findings to create an efficient prompting strategy for generating synthetic relations data for the task of Open-Vocabulary SGG. After further analysis, we showed that data generated by our method can be beneficial for the pre-training of SGG models in the Open-Vocabulary SGG, with an overall improvement of 3.5\% on the HICO-DET dataset.

Our findings mainly illustrate the current limitations of Vision-Language Models to learn fine-grained visual relationship representations \cite{yuksekgonuland}. In addition, we showed the opportunity of using synthetic data generated by VLMs for Open-Vocabulary SGG, instead of traditional captions-based approaches \cite{yuanRLIPv2FastScaling2023}.

As future work, we intend to implement a dedicated subject and object region embedding to boost the performance of Vision Language Models in relation prediction. Recent work \cite{guo2024regiongpt} has proposed a region-specific training for VLMs, such approach can be extended to the context of relation prediction with a dedicated feature refinement module for subject and object. We believe such a strategy can help the model to (1) focus on the correct regions of the image and (2) focus on the directionality of the relation.

\section*{Acknowledgments}
\label{sec:ack}

Zoe Falomir and Maëlic Neau acknowledge the Knut and Alice Wallenberg foundation and the Wallenberg AI, Autonomous Systems and Software Program (WASP).

{\small
\bibliographystyle{ieeenat_fullname}
\bibliography{11_references,OpenVoc_SGG}
}

\end{document}